\documentclass[a4paper,fleqn]{cas-sc}

\usepackage[numbers]{natbib}
\usepackage{hyperref}
\usepackage{color}
\usepackage{listings}
\usepackage{xcolor}
\usepackage{graphicx,verbatim}
\usepackage{tabularray}
\SetTblrInner{rowsep=0pt}
\usepackage{amsmath} 
\usepackage{amssymb} 

\def\tsc#1{\csdef{#1}{\textsc{\lowercase{#1}}\xspace}}
\tsc{WGM}
\tsc{QE}
\tsc{EP}
\tsc{PMS}
\tsc{BEC}
\tsc{DE}
\begin{document}
\let\WriteBookmarks\relax
\def\floatpagepagefraction{1}
\def\textpagefraction{.001}
\shorttitle{S-RRG-Bench}
\shortauthors{Yingshu Li et~al.}
\let\printorcid\relax 

\title [mode = title]{S-RRG-Bench: Structured Radiology Report Generation with Fine-Grained Evaluation Framework}                    

\author[1]{Yingshu Li}[style=chinese] 
\ead{yingshu.li@sydney.edu.au}

\author[1]{Yunyi Liu}[style=chinese] 
\ead{yunyi.liu1@sydney.edu.au} 

\author[1]{Zhanyu Wang}[style=chinese, orcid=0000-0000-0000-0000] 
\ead{zhanyu.wang@sydney.edu.au}
\ead[url]{https://wang-zhanyu.github.io/}

\author[5]{Xinyu Liang}[style=chinese] 
\ead{xinyu.liang31@gmail.com}

\author[3]{Lingqiao Liu}[style=chinese, orcid=0000-0000-0000-0000] 
\ead{lingqiao.liu@adelaide.edu.au}
\ead[URL]{https://lingqiao-adelaide.github.io/lingqiaoliu.github.io/}

\author[2]{Lei Wang}[style=chinese, orcid=0000-0000-0000-0000]
\ead{leiw@uow.edu.au}
\ead[URL]{https://sites.google.com/view/lei-hs-wang}

\author[1]{Luping Zhou}[style=chinese, orcid=0000-0000-0000-0000]
\cormark[1] 
\ead{luping.zhou@sydney.edu.au}
\ead[url]{https://sites.google.com/view/lupingzhou}
\credit{Data curation, Writing - Original draft preparation}

\address[1]{University of Sydney, New South Wales 2006, Australia}
\address[2]{University of Wollongong, New South Wales 2522, Australia}
\address[3]{University of Adelaide, South Australia 5005, Australia}
\address[4]{First Clinical Medical College, Guangzhou University of Chinese Medicine, Guangzhou 510405, China}

\cortext[1]{Corresponding author} 

\begin{abstract}
Radiology report generation (RRG) for diagnostic images, such as chest X-rays, plays a pivotal role in both clinical practice and AI. Traditional free-text reports suffer from redundancy and inconsistent language, complicating the extraction of critical clinical details. Structured radiology report generation (S-RRG) offers a promising solution by organizing information into standardized, concise formats. However, existing approaches often rely on classification or visual question answering (VQA) pipelines that require predefined label sets and produce only fragmented outputs. Template-based approaches, which generate reports by replacing keywords within fixed sentence patterns, further compromise expressiveness and often omit clinically important details. In this work, we present a novel approach to S-RRG that includes dataset construction, model training, and the introduction of a new evaluation framework. We first create a robust chest X-ray dataset (MIMIC-STRUC) that includes disease names, severity levels, probabilities, and anatomical locations, ensuring that the dataset is both clinically relevant and well-structured. We train an LLM-based model to generate standardized, high-quality reports. To assess the generated reports, we propose a specialized evaluation metric (S-Score) that not only measures disease prediction accuracy but also evaluates the precision of disease-specific details, thus offering a clinically meaningful metric for report quality that focuses on elements critical to clinical decision-making and demonstrates a stronger alignment with human assessments. Our approach highlights the effectiveness of structured reports and the importance of a tailored evaluation metric for S-RRG, providing a more clinically relevant measure of report quality.
\end{abstract}


\maketitle
\section{Introduction}

Radiology report generation (RRG)—the task of producing diagnostic descriptions from radiographs such as chest X-rays—has become a pivotal topic at the intersection of clinical practice and artificial intelligence~\citep{li2024kargen,wang2023r2gengpt,li2023comprehensive}. While most existing approaches focus on free-text generation~\citep{demner2016preparing,johnson2019mimic}, such outputs often contain redundant or stylistically inconsistent language, hindering clinicians’ ability to efficiently extract critical information~\citep{ganeshan2018structured}. Structured reporting has been proposed as a solution, offering standardized, template-based formats that enhance report clarity, consistency, and completeness~\citep{pinto2019structured,brown2019standardised}.

Recent efforts in structured radiology report generation (S-RRG) have explored classification-based and template-driven paradigms~\citep{pellegrini2023rad}, but these often constrain report expressiveness and overlook fine-grained clinical attributes such as severity, probability, and anatomical location~\citep{kale2023replace}. More recently, large language models (LLMs)~\citep{touvron2023llamaopenefficientfoundation,bai2023qwentechnicalreport} have been introduced using zero-shot or prompt-based methods~\citep{mallio2023large,pmlr-v227-keicher24a,busch2024large}; however, they suffer from limited controllability and reproducibility, with evaluation protocols still in their infancy. In this work, we aim to overcome these limitations by designing a structured dataset and evaluation metrics that more faithfully capture clinically relevant information.

\begin{figure}[h]
\centering
\centerline{\includegraphics[width=\linewidth]{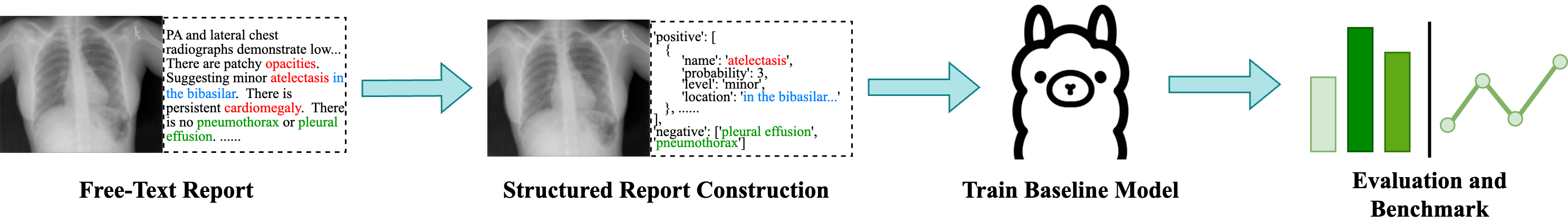}}
\caption{Overview of our proposed framework. We begin by (1) transforming free-text radiology reports into structured formats that encode key clinical attributes such as disease name, probability, severity, and anatomical location. These structured reports are represented in a machine-readable JSON format and serve as supervision signals to (2) train a generation-based baseline model. (3) Finally, we evaluate model performance using both structured metrics and natural language generation (NLG) benchmarks.}
\label{fig:overall}
\end{figure}

\textit{First, a key limitation in current research on structured radiology report generation is the absence of a comprehensive, structured dataset} that captures clinically meaningful details—such as disease name, severity, probability, and anatomical location—in a unified format. While efforts like RadReStruct~\citep{pellegrini2023rad} frame structured reporting as a hierarchical visual question answering (VQA) task, and others like Replace and Report~\citep{kale2023replace} adopt template-based classification, these approaches either fragment the report into discrete predictions or rely on insufficiently structured outputs, limiting their clinical fidelity. To address this gap, we introduce MIMIC-STRUC, a new dataset developed in collaboration with radiologists that encodes rich, clinically relevant attributes in a predefined, machine-readable JSON format. Unlike prior datasets, MIMIC-STRUC presents full structured reports rather than isolated classification results or piecemeal answers. To evaluate the utility of this format, we generate sentence-based structured reports derived directly from the JSON outputs, enabling assessment with standard NLG metrics. Empirically, we find that models trained to generate structured JSON first, followed by conversion into natural language, outperform those trained to directly produce sentence-based outputs, highlighting the benefits of structured representation for both learning efficiency and output fidelity.

\textit{Second, there is a lack of specialized evaluation metrics for structured radiology reports.} 
Clinical metrics like Radgraph F1~\citep{jain2021radgraph}, Chexbert F1~\citep{smit2020chexbert}, and Radcliq~\citep{yu2023evaluating} require additional training and may be prone to errors when extracting medically relevant content, potentially overlooking important clinical details. To address this, we propose the \textbf{Structured Report Score (S-Score)}, which evaluates disease prediction accuracy and the precision of key details. We validate our metric by calculating its correlation with GPT scores, as studies~\citep{liu2024mrscore,liu2024er2score} demonstrate that GPT-based evaluations align well with radiologist assessments, showing strong alignment and confirming its clinical relevance.

\textit{Third, there is currently no established baseline model for generating fully structured radiology reports.} Existing structured report generation methods often rely on classification or hierarchical VQA frameworks to predict specific attributes at different levels~\citep{pellegrini2023rad}, requiring predefined label sets and hand-crafted templates. In contrast, we propose a generation-based approach that directly produces the entire structured report in a single step, without depending on predefined labels. Inspired by recent advances demonstrating the ability of large language models (LLMs) to generate well-formed structured outputs such as JSON~\citep{tang2023struc}, we train an LLM-based model for structured radiology report generation (S-RRG).

Our contributions are summarized as follows and illustrated in Fig.\ref{fig:overall}:
\begin{itemize}
    \item[(1)] We construct a new chest X-ray structured report generation dataset, which includes disease names, severity levels, probabilities, and locations, and demonstrate the efficacy of our proposed structured format.
    \item[(2)] We propose the Structured Report Score (S-Score), a new evaluation metric that assesses both disease prediction accuracy and the precision of clinical details. This metric outperforms traditional metrics (e.g., BLEU, ROUGE) and aligns well with radiologist assessments, as demonstrated by its strong Kendall Tau and Spearman correlation with human-like GPT evaluations.
    \item[(3)] We establish a baseline model for structured report generation, comparing traditional transformer-based methods with LLM-based models. Our results demonstrate that LLMs significantly improve accuracy and feasibility, and we propose a new benchmarking task for evaluation.
\end{itemize}

\section{Related Works}
\subsection{Radiology Report Generation}
Radiology report generation (RRG) aims to automatically translate medical images, such as chest X-rays into coherent diagnostic reports. An effective radiology report must not only mention the correct diseases but also accurately describe their clinical attributes such as severity, location, and uncertainty~\citep{hartung2020create}. To this end, early studies employed encoder-decoder architectures, such as memory-driven transformers designed to retain critical information throughout the generation process~\cite{chen2022cross}. METransformer~\citep{wang2023metransformer} advanced this line of work by introducing multiple learnable “expert” tokens into both the encoder and decoder of a Transformer architecture. These tokens enable the model to capture diverse disease-specific patterns and facilitate cross-modal alignment between image regions and textual descriptions.

More recently, large language models (LLMs) have been applied to radiology report generation. R2GenGPT~\citep{wang2023r2gengpt}, for example, integrates a Swin Transformer as the image encoder and LLaMA as the language model, leveraging medical prompts to enhance disease awareness during generation. MAIRA-2~\citep{bannur2024maira} introduces a multi-agent collaborative framework with domain-specific instruction tuning, demonstrating strong clinical performance across multiple benchmarks.

\subsection{Structured Radiology Report Generation}
Structured radiology report generation (S-RRG) aims to transform radiological findings into standardized formats that enhance interpretability, clinical utility, and readability. This structured approach addresses limitations of conventional free-text reports, including redundancy, linguistic variability, and challenges in automated analysis.

Recent approaches to S-RRG typically relied on classification-based methods. For example, RadReStruct~\citep{pellegrini2023rad} formulates structured reporting as a hierarchical visual question answering (VQA) task, utilizing an autoregressive framework to populate structured report fields through a sequence of image-grounded questions. Similarly, FlexR~\citep{pmlr-v227-keicher24a} treats S-RRG as a few-shot classification task, representing each structured sentence (e.g., “There is mild cardiomegaly”) as a distinct class, with predictions generated by aligning image features with language embeddings via contrastive learning~\citep{zhai2023sigmoid}. Replace and Report~\citep{kale2023replace} takes a template-based route, first designing a free-text report template, then replacing pre-defined slots within semi-structured information. Yet its templates remain largely narrative and omit critical attributes such as probability, severity, and location details that are indispensable for clinical decision-making. Extending these paradigms, \citet{zhang2024new} introduced a large-scale chest X-ray benchmark enriched with clinically relevant annotations, such as uncertainty, severity, and localization-aware labels, thereby advancing beyond conventional binary classification. While effective within classification frameworks, these methods often constrain the expressiveness and flexibility of the generated reports, and typically rely on hand-crafted templates or predefined label sets. In parallel, recent research has investigated the application of large language models (LLMs) to structured radiology reporting~\citep{busch2024large,mallio2023large}. However, most existing LLM-based approaches adopt zero-shot or prompt-based methods to either convert free-text reports into structured formats or generate free-text from structured inputs. These techniques are limited by reduced controllability, poor reproducibility, and underdeveloped evaluation protocols.

To address these limitations, we frame structured radiology report generation as a supervised, generation-based task that directly predicts fully structured reports in JSON format, eliminating the need for post-hoc classification or retrieval-based pipelines. Unlike classification-based methods or prompt-tuned LLMs, our approach facilitates greater expressiveness and fine-grained control over entity-level attributes, including disease name, probability, severity, and anatomical location. Furthermore, we introduce MIMIC-STRUC, a new benchmark that includes detailed structured annotations designed to support comprehensive evaluation of structured report generation.

\subsection{Evaluation Metrics for Radiology Report Generation}
Assessing the quality of generated radiology reports remains a significant open challenge. Conventional natural language metrics such as BLEU and ROUGE primarily capture surface-level n-gram overlap and fail to account for clinically salient information. While domain-specific metrics have been developed, each presents some limitations. For instance, RadGraph F1~\citep{jain2021radgraph} and CheXbert F1~\citep{smit2020chexbert} rely on auxiliary information extraction models, whose inherent errors can propagate into the final metric. RadCliQ~\citep{yu2023evaluating} aggregates multiple clinical signals but cannot disentangle whether discrepancies arise from incorrect disease classification or from missing fine-grained attributes. Recent studies have explored leveraging large language models (LLMs) to approximate radiologist judgments, either by finetuning LLM-based evaluators~\citep{liu2024mrscore,liu2024er2score} or employing zero-shot and few-shot prompting. However, these methods demand considerable computational resources. To enable more faithful and interpretable evaluation, we propose the Structured-Report Score (S-Score), which decomposes performance into (i) disease prediction accuracy and (ii) detail precision across severity, probability, and anatomical location.
\section{Methodology}
\begin{figure}[h]
\centering
\centerline{\includegraphics[width=\linewidth]{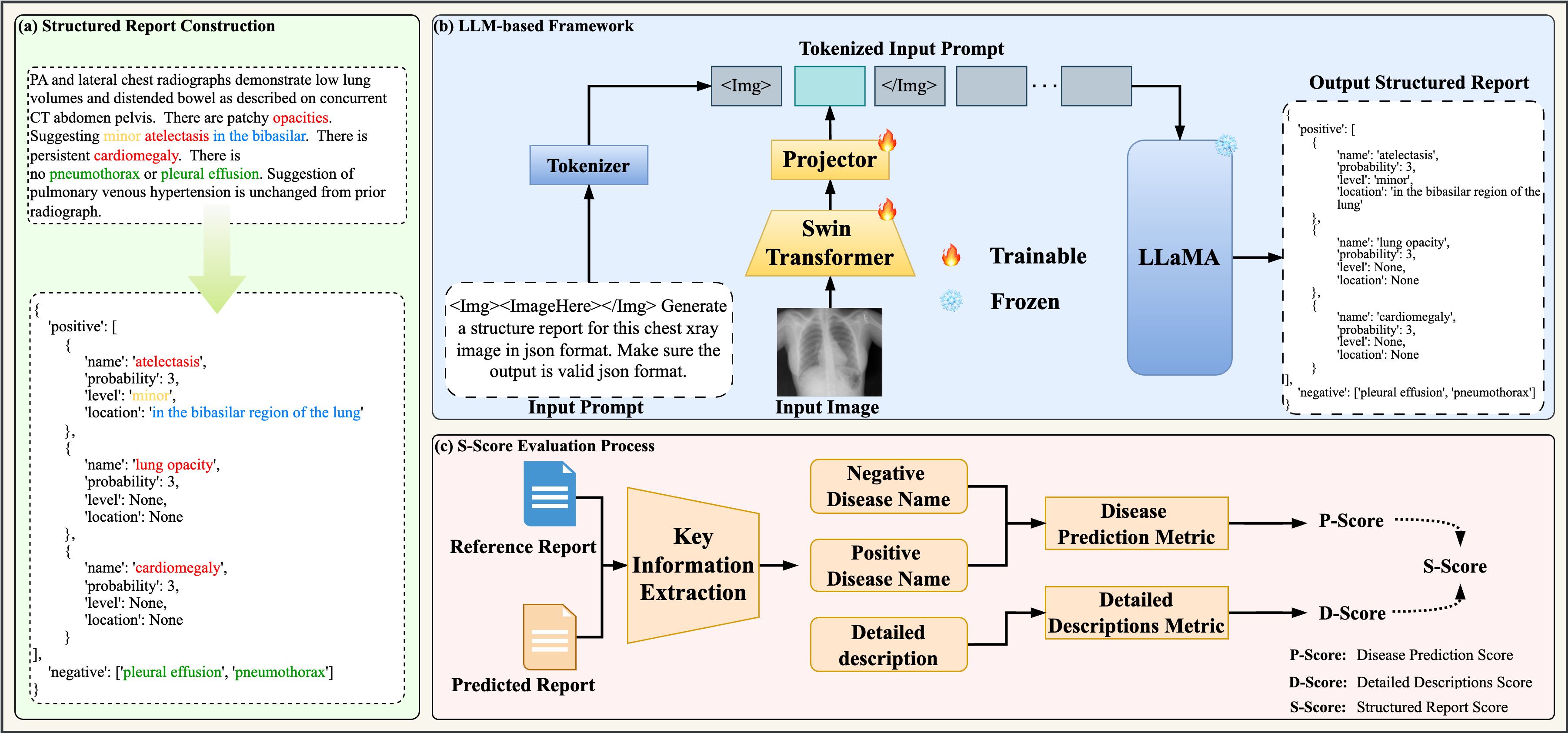}}
\caption{Overview of our framework. (a) Construction of a structured report from a radiology report, where key information is extracted and represented in JSON format, with key elements highlighted in different colors. (b) The LLM-based framework. (c) The evaluation process, extracting key information from both references and predicted reports to compute the evaluation metrics: P-Score (Disease Prediction), D-Score (Detailed Descriptions), and S-Score (Structured Report Score).}
\label{fig:main}
\end{figure}

As illustrated in Fig.\ref{fig:main}, our proposed framework for structured radiology report generation consists of three main components: (a) structured report construction, (b) a generation-based modeling framework, and (c) an evaluation pipeline.

\subsection{Structured Reports Dataset Construction}
As highlighted by~\citet{hartung2020create}, an effective radiology report should support rapid access to both disease-related information and its associated clinical details. However, free-text reports often contain redundant language and variable phrasing, making automated interpretation and structured evaluation challenging. To overcome this, we transform them into a machine-readable \textbf{JSON format} (Fig.~\ref{fig:main}(a)), which (1) structures findings as clear key-value pairs, (2) removes unnecessary linguistic noise, and (3) enables efficient extraction and evaluation, making it ideal for supervised training and structured metrics. A complete structured report includes the disease name for both positive and negative findings. For each positive finding, additional attributes are recorded to capture clinical nuance: (i) probability, reflecting diagnostic certainty (e.g., “may,” “might,” “possible”); (ii) severity level, indicating disease intensity (e.g., “mild,” “moderate,” “severe”); and (iii) anatomical location, specifying the affected region (e.g., “upper lobe,” “right lung”). These components collectively define the structure of a clinically meaningful report~\citep{zhang2024new}. We incorporate several curated keyword lists (e.g., disease names and location terms) provided by \citep{zhang2024new} to extract the clinic-related information. We utilize the \textsc{NLTK}\footnote{\url{https://www.nltk.org/}} and \textsc{spaCy}\footnote{\url{https://spacy.io/}} libraries for biomedical text preprocessing and entity recognition from free-text radiology reports.

\noindent\textbf{Disease Name Extraction}~~
To support a broader diagnostic spectrum, we extend the disease vocabulary beyond the 14 CheXbert categories~\citep{smit2020chexbert} to include 30 disease types curated in \citep{zhang2024new}. Reports are tokenized and parsed using NLTK and spaCy, and disease mentions are matched against this extended list to generate structured disease name annotations.

\noindent\textbf{Probability Extraction}~~
Diagnostic probability is inferred from hedging expressions such as “may”, “might”, and “could”. These cues are matched against an uncertainty keyword list and mapped to discrete values in $\{1, 2, 3\}$, with higher scores indicating greater diagnostic certainty.

\noindent\textbf{Severity Level Extraction}~~
We extract severity indicators such as “mild”, “moderate”, and “severe”, and link them to their corresponding disease mentions based on proximity and sentence structure, ensuring accurate context-dependent labeling.

\begin{figure*}[h]
\centering
\centerline{\includegraphics[width=\linewidth]{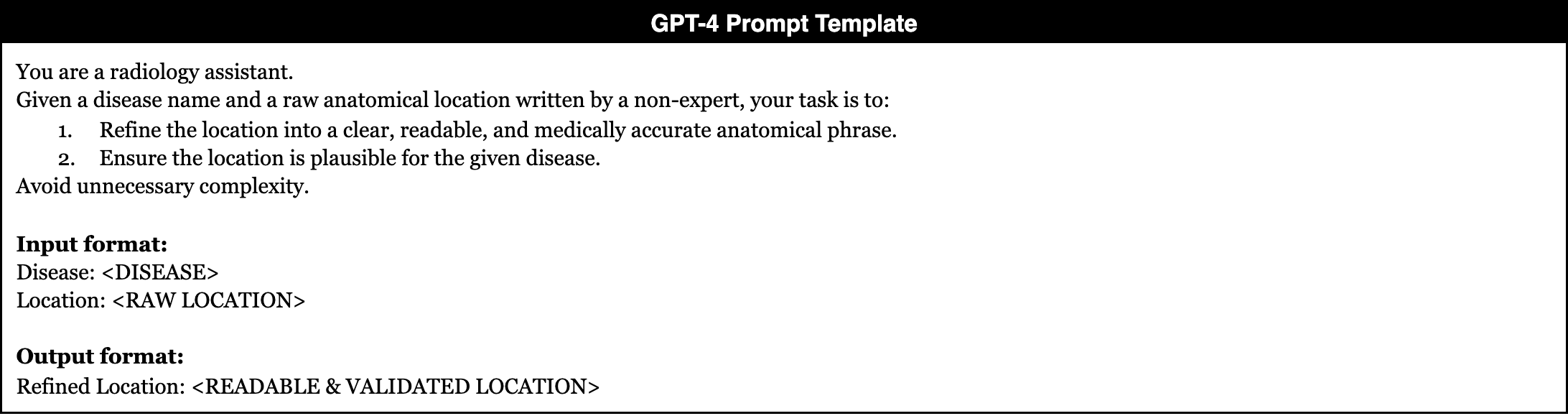}}
\caption{This prompt guides GPT-4 to refine raw anatomical location phrases provided by non-expert annotators.}
\label{fig:loca}
\end{figure*}

\noindent\textbf{Location Extraction}~~
Anatomical locations (e.g., “upper lobe”, “right lung”) are identified to indicate where the disease manifests. However, the extracted phrases are often vague or syntactically ambiguous. To improve clinical clarity and consistency, we employ GPT-4~\citep{achiam2023gpt} to rephrase these descriptions into standardized, clinician-friendly expressions (e.g., “lung right lower” $\rightarrow$ “in the lower zone of the right lung”). The detailed prompt is in Fig~\ref{fig:loca}.

The final structured report template is in Fig.\ref{fig:template}:
\begin{figure}[h]
\centering
\centerline{\includegraphics[width=0.4\linewidth]{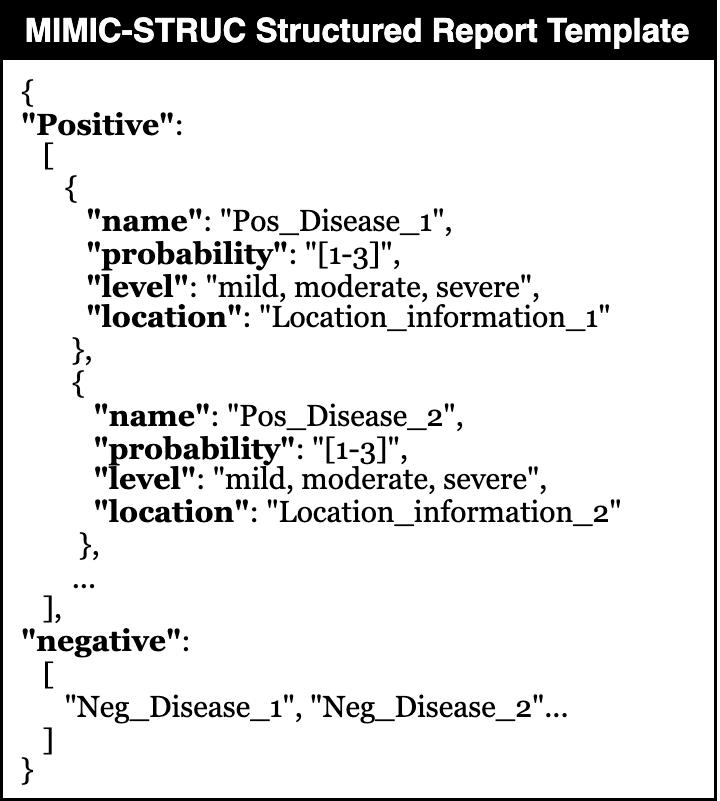}}
\caption{MIMIC-STRUC Structured Report Template: The template defines the standardized JSON format used in MIMIC-STRUC, where each positive finding includes four clinically relevant attributes—name, probability, level (severity), and location—while negative findings are listed by name. }
\label{fig:template}
\end{figure}

\begin{figure*}[h]
\centering
\centerline{\includegraphics[width=\linewidth]{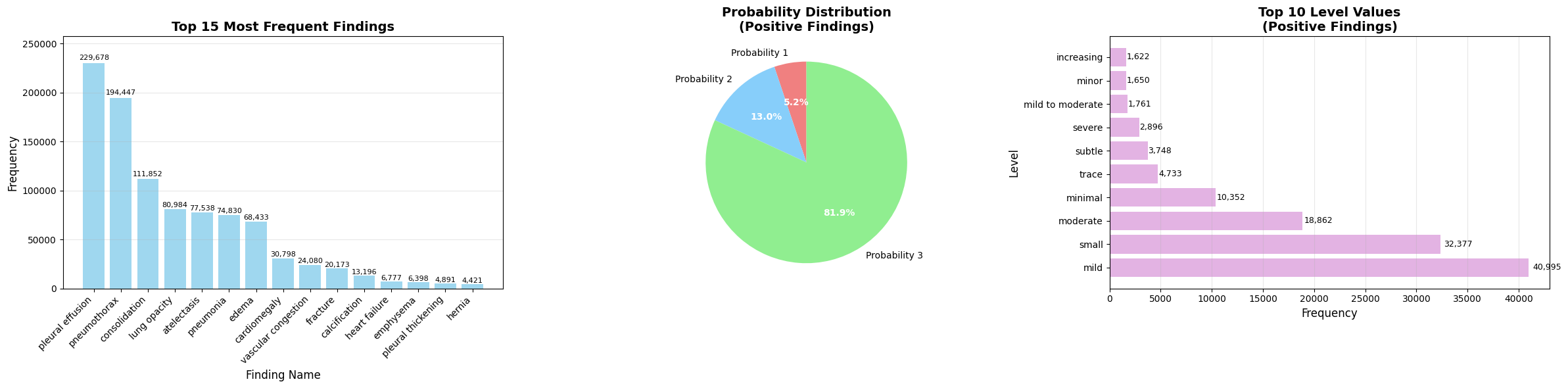}}
\caption{Summary statistics and visualizations of MIMIC-STRUC}
\label{fig:vis}
\end{figure*}

To further illustrate the distribution and richness of our structured annotations, we present summary statistics of the extracted attributes in Figure~\ref{fig:vis}. The left panel shows the most frequently identified disease types, while the middle and right panels visualize the distributions of probability scores and severity levels for positive findings, respectively.

\subsection{S-Score: An Evaluation Metric for Structured Reports}
Since S-RRG is an underexplored task, no established metric exists to evaluate the quality of generated reports. Inspired by clinical evaluation scores like Chexbert F1~\citep{smit2020chexbert}, Radgraph F1~\citep{jain2021radgraph}, and RadcliQ~\citep{yu2023evaluating}, we introduce the S-Score (Structured Report Score), a comprehensive metric designed to assess both disease prediction accuracy and the precision of detailed descriptions. Unlike traditional clinical scores that require additional models for information extraction, potentially introducing errors, our metric leverages the structured format, ensuring more accurate extraction. For example, CheXbert F1 relies on a sentence-level classifier to map reports to 14 disease labels, which may misinterpret negation, uncertainty, or co-occurring conditions. RadGraph F1 depends on named entity recognition and relation extraction, which often suffer from span detection and linking errors. In contrast, S-Score directly compares model outputs in structured JSON form, avoiding such error-prone intermediate steps and improving reliability. Moreover, both CheXbert and RadGraph F1 are designed and trained specifically for free-text radiology reports; their applicability to structured report generation remains limited. As structured reporting becomes increasingly adopted, there is a growing need for dedicated metrics like S-Score that can more faithfully evaluate clinical completeness and precision in structured formats.
Additionally, the S-Score evaluates data at multiple granularities, providing a more thorough assessment. For instance, Chexbert F1 evaluates only 14 predefined diseases, whereas the S-Score considered all the diseases in report. also considers detailed aspects like disease location.

\noindent\textbf{Disease Prediction Score}~~In medical image analysis, the accuracy of disease prediction plays a crucial role in determining the effectiveness of a model. To better evaluate the accuracy of diseases prediction, we introduce a Disease Prediction Score that calculates the precision and recall for both positive and negative disease predictions. For positive disease predictions, we define one reference set of disease names as $\text{Ref}_{pos}^{i} = {(\omega_{r1},...,\omega_{rN})}$, where $\omega_{rN}$ represents the name of a predicted disease, and one hypothesis set as $\text{Hyp}_{pos}^{i} = {(\omega_{h1},...,\omega_{hM})}$, where $\omega_{hM}$ represents the name of a ground truth disease name. The combined set $\Omega$ of both reference and hypothesis disease names is: $\Omega^{i} = \text{Ref}_{pos}^{i} \cup \text{Hyp}_{pos}^{i}$. For each disease $\omega$ in $\Omega^i$, the true label and predicted label are:

\begin{flalign}
&&
y_{\text{true}}^{i}(\omega) = 
\begin{cases}
1, & \text{if } \omega \in \text{Ref}_{\text{pos}}^{i}, \\
0, & \text{otherwise}.
\end{cases}
&&
\end{flalign}

\begin{flalign} &&
y_{\text{pred}}^{i}(\omega) = 
\begin{cases}
1, & \text{if } \omega \in \text{Hyp}_{\text{pos}}^{i}, \\
0, & \text{otherwise}.
\end{cases}
&& \end{flalign}

The Disease prediction score for positive diseases can be obtained by:
\begin{flalign} &&
    \text{P}\_\text{Score}_{pos}^{i} = \frac{2 \times \text{Precision}_{pos} \times \text{Recall}_{pos}}{\text{Precision}_{pos} + \text{Recall}_{pos}}
&& \end{flalign}

\begin{flalign} &&
    \text{Precision}_{pos}^{i} = \frac{\sum_{\omega \in \Omega} (y_{true}(\omega) \cap y_{pred}(\omega))}{\sum_{\omega \in \Omega} y_{pred}(\omega)}
&& \end{flalign}

\begin{flalign} &&
    \text{Recall}_{pos}^{i} = \frac{\sum_{\omega \in \Omega} (y_{true}(\omega) \cap y_{pred}(\omega))}{\sum_{\omega \in \Omega} y_{true}(\omega)}
&& \end{flalign}

A similar formulation applies to the disease prediction score for negative diseases. The overall disease prediction score is:
\begin{flalign} &&
    \text{P}\_\text{Score}^{i} = \frac{1}{2}(\text{P}\_\text{Score}_{pos}^{i} + \text{P}\_\text{Score}_{neg}^{i})
&& \end{flalign}

\noindent\textbf{Detailed Descriptions Score}~~This Score is designed to assess the accuracy of a model's predictions for fine-grained details such as severity level, probability, and location in radiology reports. Unlike traditional metrics, which assess overall report quality, this score evaluates three aspects: level, probability, and location. To compute it, we first check if the disease name matches between the reference and the prediction. The detailed descriptions are represented as: $\{prob_i, level_i, loc_i\}$ for the reference and $\{prob_i', level_i', loc_i'\}$ for the prediction.

\noindent \underline{Probability Score}~~For each matching pair $i$, the probability score $s_{prob}(i)$ is:
\begin{flalign} &&
s_{\text{prob}}^{i} =
\begin{cases}
1 - \text{MSE}(prob_i, prob_i'), & \text{if } prob_i' \text{ is predicted}, \\
0, & \text{otherwise}.
\end{cases}
&& \end{flalign}
\noindent \underline{Level Score}~~For each matching pair $i$, the level score $s_{level}(i)$ is calculated as:
\begin{flalign} &&
s_{\text{level}}^{i} =
\begin{cases}
1, & \text{if } level_i = level_i', \\
0, & \text{otherwise}.
\end{cases}
&& \end{flalign}

\noindent \underline{Location Score}~~The Location Score evaluates the accuracy of predicted location descriptions in the radiology report. Since location descriptions are typically short phrases, BLEU~\citep{papineni2002bleu} is suitable here as it effectively captures exact word-level matches in a short text, providing an accurate measure of location prediction.
\begin{flalign} &&
s_{\text{loc}}^{i} =
\begin{cases}
\text{BLEU}(loc_i, loc_i'), & \text{if } loc_i' \text{ is predicted}, \\
0, & \text{otherwise}.
\end{cases}
&& \end{flalign}

The final detailed description score can be formulated as: 
\begin{flalign} &&
\text{D}\_\text{Score}^{i} = 
\mathbf{1}_{\text{name\_match}} \left( w_{\text{prob}} \cdot s_{\text{prob}}^{i} + w_{\text{level}} \cdot s_{\text{level}}^{i} + w_{\text{loc}} \cdot s_{\text{loc}}^{i} \right).
&& \end{flalign}
Here, $w_{\text{prob}}$, $w_{\text{level}}$, and $w_{\text{loc}}$ are used to balance the score. The binary indicator $\mathbf{1}_{\text{name\_match}}$ ensures that scores for level, probability, and location are computed only when the name matches; otherwise, the score for that pair is set to 0. In our implementation, we assign equal weights of $\frac{1}{3}$ to probability, severity, and location, reflecting their equal importance in clinical interpretation. The final S-Score is:
\begin{flalign} &&
    \text{S-Score}^{i} = \frac{1}{2}(\text{P}\_\text{Score}^{i} + \text{D}\_\text{Score}^{i})
&& \end{flalign}

\subsection{Structured Report Generation Baseline}
Large language models (LLMs) have recently demonstrated remarkable capabilities and robustness in generating structured data, such as JSON and HTML formats, further reinforcing the feasibility of generating structured radiology reports~\citep{tang2023struc}. As illustrated in Fig.\ref{fig:main}(b), our model consists of three main components: a visual encoder, a projector, and an LLM. 

\noindent\textbf{Visual Feature Extraction}~~For an input chest X-ray image $X_v$, we used the Swin Transformer as visual encoder~\citep{liu2021swin}, which produces the visual features $Z_v = \text{Swin}(X_v)$. The features from the last transformer layer are used in our work. To connect the image features to the LLM's word feature space, we employ a two-layer MLP as the projector. Specifically, the process is $H_v = \text{MLP}(Z_v)$.

\noindent\textbf{Structured Report Generation}~~For report generation, we adopt the LLaMA3-3B model. Although the 3B model is a lightweight LLM, it still demonstrates strong capabilities in generating structured data. Thus, we have selected it for this task to balance performance and computational efficiency. Given the extracted visual features $H_v$, our prompt is designed as: \textit{$<|start\_header\_id|>user<|end\_header\_id|>$ <Img>$H_v$</Img>, $X_p$. $<|eot\_id|><|start\_header\_id|>assistant<|end\_header\_id|>$}

Here $\mathbf{X}_p$ is our designed instruction prompt specific to the task. In our current implementation, $\mathbf{X}_p$ = "Generate a structure report for this chest x-ray image in JSON format. Make sure the output is valid JSON format.". 
Our overall model is optimized by minimizing the cross-entropy loss:
\begin{flalign} &&
\mathcal{L}_{CE}(\theta) = -\sum_{i=1}^{N_r} \log p_\theta (t_i^{*}|X_v, X_P,t_{1:i-1}^{*}),
&& \end{flalign}
where $\theta$ denotes model parameters, and $t_i^{*}$ is the $i$-th word in the ground truth report with a length of $N_r$ words.

\section{Experiments}
\subsection{Experimental Settings}
\noindent\textbf{Dataset}~~We propose MIMIC-STRUC, derived from the MIMIC-CXR database~\citep{johnson2019mimic}, the largest publicly available chest X-ray collection, which contains 377,110 images and 227,835 reports from 64,588 patients. We extracted key clinical information (disease names, severity levels, and locations) and converted this data into a structured JSON format. This format retains only the relevant information, providing a concise and efficient representation of the essential elements. Our dataset includes 257,165 images for training and 3,735 images for testing.

\noindent\textbf{Implementation Details}~~Our baseline model employs LLaMA3-3B\footnote{\url{https://huggingface.co/meta-llama/Llama-3.2-3B}} as the large language model (LLM), which is kept frozen during training, and a Swin Transformer\footnote{\url{https://huggingface.co/microsoft/swin-base-patch4-window7-224}} as the visual encoder, which is jointly optimized. Training is performed using mixed precision on four NVIDIA A6000 48GB GPUs for 3 epochs, with a mini-batch size of 8 and a learning rate of 1e-4. For inference, we use beam search with a beam size of 3. To evaluate structured outputs, we load both predicted and reference reports using \texttt{simplejson}\footnote{\url{https://pypi.org/project/simplejson/}}. Before parsing, we normalize output strings by correcting invalid tokens and fixing formatting issues using \texttt{json-repair}\footnote{\url{https://github.com/mangiucugna/json_repair}}. After loading, we extract structured fields from both \texttt{positive} and \texttt{negative} sections—namely: disease name, probability, severity level, and location. These fields are used to compute our fine-grained evaluation sub-scores.

\subsection{Results and Discussion}
\noindent \textbf{Efficacy of LLMs}~~
To assess the impact of model architecture, we compare the LLaMA-based setup with a transformer-based decoder. Specifically, we replace the LLM with a 6-layer Transformer decoder while keeping the image encoder fixed; the Transformer decoder is randomly initialized and trained end-to-end during optimization. As shown in Table~\ref{tab2}, the ability of LLMs to handle structured outputs significantly enhances performance in structured radiology report generation. From LLaMA3-3B to LLaMA3-8B, we observe consistent improvements across all metrics, indicating that scaling up LLMs improves their capacity to generate accurate and well-structured reports. Furthermore, newer lightweight models such as Phi-3 (4B)~\cite{abdin2024phi} and Qwen2.5 (3B)~\cite{bai2023qwentechnicalreport} achieve even better performance despite having fewer parameters, further validating that LLMs possess the capabilities necessary for effective structured radiology report generation. These findings demonstrate that LLMs are well-suited for end-to-end structured radiology report generation, moving beyond classification or template-based methods toward more flexible, unified modeling.


\begin{table}[h]
\centering
\caption{Comparison with different methods. The highest scores are in bold.}
\label{tab2}
\resizebox{\linewidth}{!}{
\begin{tabular}{l|cccc|ccc}
\hline
Methods                     & BLEU-1      & BLEU-4         & METEOR       & ROUGE       & P-Score      & D-Score      & S-Score  \\  \hline
Transformer Based           & 0.495       & 0.268          & 0.382        & 0.537       & 0.444        & 0.344        & 0.394    \\  \hline
LLaMA3 3B                   & 0.560       & 0.399          & 0.465        & 0.552       & 0.459        & 0.377        & 0.418    \\  \hline
Phi3 4B                     & 0.583       & 0.415          & 0.464        & 0.558       & 0.465        & 0.418        & 0.442    \\  \hline
Qwen2.5 3B                  & 0.580       & 0.415          & \textbf{0.471}             & 0.561        & 0.466        & \textbf{0.427 }       & \textbf{0.446}    \\  \hline
LLaMA3 8B                   & \textbf{0.602}       & \textbf{0.425}          & 0.463        & \textbf{0.561}       & \textbf{0.468}        & 0.405        & \textbf{0.436}    \\  \hline
\end{tabular}
}
\end{table}

\noindent \textbf{Efficacy of JSON Format}~~To assess the representational effectiveness of the JSON-based format, we designed an experiment comparing two generation strategies under a unified evaluation setup. Importantly, the ground-truth references used in this experiment are sentence-style reports derived from our structured JSON annotations—not the original free-text MIMIC reports. These sentence-style reports were developed in collaboration with board-certified radiologists (each with over 10 years of experience) to ensure clinical fidelity and preservation of key attributes such as disease name, probability, severity, and anatomical location.

The transformation templates are as follows:
\begin{flushleft}
\textbf{For positive disease:} there \colorbox{gray!30}{\textit{probability}} be \colorbox{gray!30}{\textit{level}} \colorbox{gray!30}{\textit{name}} \colorbox{gray!30}{\textit{location}}. \\
\textbf{For negative disease:} no evidence of \colorbox{gray!30}{\textit{name}}.
\end{flushleft}

We compare two generation strategies: \underline{Structured-First}, where the model first generates a structured report in JSON format, which is then converted into sentence-style text using predefined templates, and \underline{Sentence-Direct}, where the model is trained to generate the sentence-style reports directly. As shown in Table~\ref{tab1}, the Structured-First approach yields higher BLEU and METEOR scores, suggesting that generating structured representations first leads to more accurate and complete report outputs under our evaluation protocol.

While standard NLG metrics such as BLEU and METEOR have limitations in evaluating free-text clinical reports—particularly regarding semantic and diagnostic accuracy—they are appropriate in our controlled setting. Given that the sentence templates are fixed and variation is limited to key clinical attributes (e.g., disease name, severity, probability, and location), these metrics provide a direct and interpretable measure of how accurately the generated content reflects these structured elements. As such, they offer a practical and consistent means of assessing the fidelity of generated outputs in this specific task formulation.

\begin{table}[h!]
\centering
\caption{Comparison of Structured-First and Sentence-Direct S-RRG.}\label{tab1}
\resizebox{\linewidth}{!}{
\begin{tabular}{l|ccccccccc}
\hline
Methods                                                 & BLEU-1      & BLEU-2         & BLEU-3      & BLEU-4         & METEOR       & ROUGE        \\ \hline
Sentence-Direct                                          & 0.544       & 0.483          & 0.425       & 0.377          & 0.297        & \textbf{0.540}        \\ \hline
Structured-First                                        & \textbf{0.575}       & \textbf{0.506}          & \textbf{0.439}       & \textbf{0.383}         & \textbf{0.300}        & 0.521        \\ \hline
\end{tabular}
}
\end{table}

\begin{figure*}[h]
\centering
\centerline{\includegraphics[width=\linewidth]{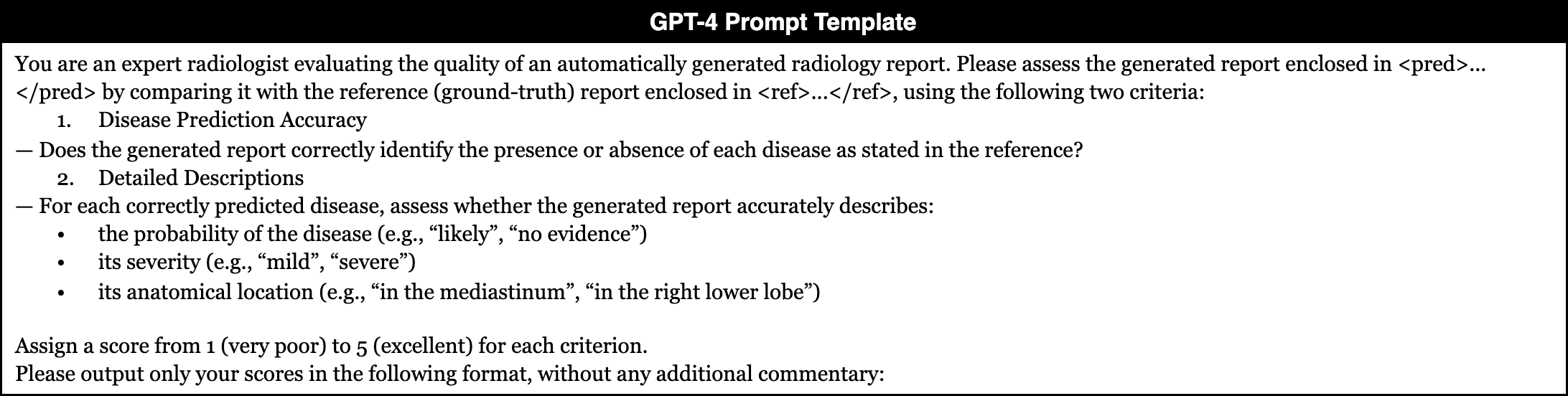}}
\caption{GPT-4 evaluation prompt used to validate S-Score. The prompt instructs GPT-4 to evaluate generated reports along two clinically relevant dimensions—Disease Prediction and Detailed Descriptions—based on comparison with ground-truth reports. This structure mirrors the design of S-Score and ensures alignment with human judgment.}
\label{fig:gpt}
\end{figure*}

\noindent \textbf{Validation of S-Score}~~To assess the reliability of the proposed S-Score and address the impracticality of extensive manual review, we employ GPT-4 as a proxy evaluator to determine how effectively S-Score captures clinically meaningful quality signals. We conducted a correlation analysis using GPT-4, which prior studies, such as MRScore~\cite{liu2024mrscore} and RaTE~\cite{zhao2024ratescore}, have demonstrated to approximate expert-level evaluations and closely align with human judgment in clinical text assessment tasks~\citep{liu2024mrscore,zhao2024ratescore}. Accordingly, we instructed GPT-4 to evaluate generated reports along two key dimensions—\textit{Disease Prediction} and \textit{Detailed Descriptions}—which directly correspond to the structure of S-Score. The detailed prompt is in Fig.\ref{fig:gpt}. As shown in Table~\ref{tab3}, we present the Kendall Tau and Spearman correlations between GPT-4 ratings and several evaluation metrics, including BLEU, ROUGE, METEOR, GREEN~\cite{ostmeier2024green}, and our proposed S-Score. Among all metrics, S-Score achieves the highest correlation with GPT-4, with a Kendall Tau of 0.591 and a Spearman correlation of 0.769. Notably, while GREEN, a recent LLM-based evaluation metric, performs competitively (Kendall Tau: 0.444; Spearman: 0.572), it still falls short of S-Score. These results indicate that S-Score aligns more closely with human evaluations than both traditional NLG metrics (e.g., BLEU-4: 0.392/0.541) and LLM-based alternatives. Moreover, unlike GPT and GREEN, S-Score is lightweight, fully interpretable, and does not rely on any external language model at inference time—making it a more practical and scalable solution for structured radiology report evaluation in real-world clinical settings.

\begin{table}[h!]
\centering
\caption{Correlation between evaluation metrics and GPT-4’s human-like scoring.}\label{tab3}
\resizebox{\linewidth}{!}{
\begin{tabular}{l|ccccccccc}
\hline
Evaluation Metrics                               & BLEU-1       & BLEU-2        & BLEU-3        & BLEU-4      & METEOR         & ROUGE    & GREEN    & S-Score               \\ \hline
Kendall Tau                                      & 0.267        & 0.323         & 0.363         & 0.392       & 0.221          & 0.368    & 0.444    & \textbf{0.591}        \\ \hline
Spearman                                         & 0.381        & 0.456         & 0.506         & 0.541       & 0.314          & 0.518    & 0.572    & \textbf{0.769}        \\ \hline
\end{tabular}
}
\end{table}

\begin{figure*}[h]
\centering
\centerline{\includegraphics[width=\linewidth]{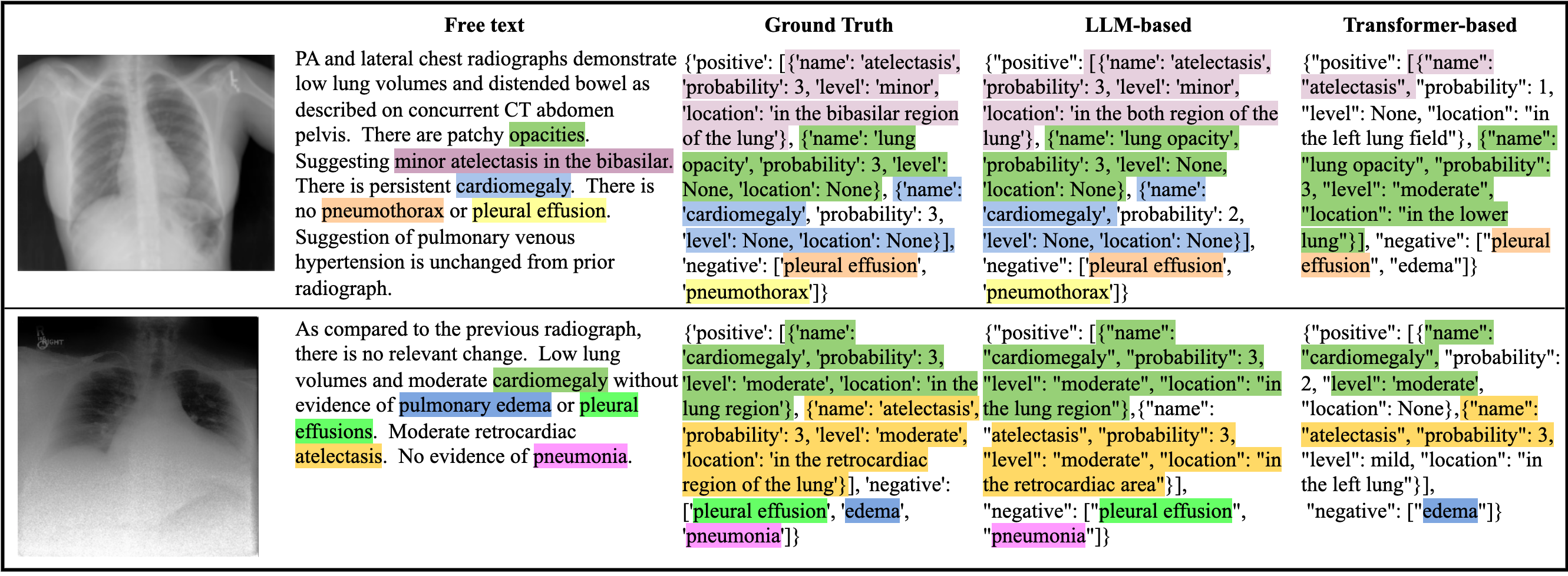}}
\caption{Examples of the generated reports. For better illustration, different colours
highlight different medical terms in the reports.}
\label{fig:case}
\end{figure*}

\noindent \textbf{Qualitative Results.}~~In Fig.\ref{fig:case}, we present a qualitative comparison between the structured reports generated by the LLM-based model and the Transformer-based model, alongside the ground truth annotations. To enhance interpretability, we highlight key clinical entities using consistent color across representations.

In the first example (top row), both models correctly extract major findings such as atelectasis, lung opacity, and cardiomegaly, with the LLM-based output showing better alignment in probability and location attributes. In contrast, the Transformer-based model underestimates the probability of atelectasis and fails to provide complete severity or location information for cardiomegaly, reflecting weaker detail fidelity.

In the second example (bottom row), the LLM-based model again closely matches the ground truth, correctly identifying both cardiomegaly and atelectasis, and providing clinically coherent descriptions. However, it misses the negative mention of edema. The Transformer-based model further underperforms by omitting pneumonia from the negatives and simplifying severity and location fields. These examples illustrate that while both models can recognize major findings, the LLM-based approach produces more accurate and complete structured reports, particularly in capturing fine-grained clinical attributes.

\section{Conclusions}
In this paper, we introduce a comprehensive structured radiology report dataset, including details like disease severity, location, and probability, addressing gaps in existing datasets. We also propose the S-Score, an evaluation metric for assessing both prediction accuracy and the precision of report details. Our experiments show that this dataset and metric offer a more clinically relevant and accurate assessment, establishing a new benchmark for structured radiology report generation.

\section{Declaration of generative AI and AI-assisted technologies in the writing process}

During the preparation of this work the author(s) used ChatGPT in order to assist with language refinement. After using this tool/service, the author(s) reviewed and edited the content as needed and take(s) full responsibility for the content of the publication.



\bibliographystyle{cas-model2-names}

\bibliography{cas-refs}






\end{document}